\documentclass[letterpaper]{article}

\usepackage{natbib}
\usepackage{alifeconf}
\usepackage{algorithm}
\usepackage[noend]{algpseudocode}
\usepackage{amsmath,amssymb,amsthm}
\usepackage{booktabs}
\usepackage{enumitem}
\usepackage{graphicx}
\usepackage[hidelinks]{hyperref} 
\usepackage{cleveref}
\usepackage{multirow}
\usepackage[autolanguage]{numprint} \npthousandsep{,}
\usepackage{pifont}  
\usepackage{tikz}
\usepackage{subcaption}
\usepackage{tablefootnote}

\newcommand{\Gtri}{\ensuremath{G_\Delta}}
\newcommand{\alg}{\ensuremath{\mathcal{A}}}
\newcommand{\numparticles}{\ensuremath{n}}
\newcommand{\numcolors}{\ensuremath{c}}
\newcommand{\density}{\ensuremath{\delta}}
\newcommand{\sopssizes}{\ensuremath{S}}
\newcommand{\numtrials}{\ensuremath{T}}
\newcommand{\population}{\ensuremath{P}}
\newcommand{\numgens}{\ensuremath{G}}
\newcommand{\mutrate}{\ensuremath{M}}
\newcommand{\hypermut}{\ensuremath{H}}
\newcommand{\evosops}{\textsc{EvoSOPS}}

\definecolor{color1}{RGB}{24, 85, 98}
\definecolor{color2}{RGB}{87, 118, 71}
\definecolor{color3}{RGB}{179, 142, 137}
\definecolor{backclr}{RGB}{188, 188, 188}

\makeatletter
\newif\ifanon
\anonfalse  
\makeatother

\newif\ifcomment
\commenttrue     

\newif\iffigabbrv
\figabbrvfalse   
\newcommand{\figtext}{\iffigabbrv Fig.\else Figure\fi}

\newif\ifeqabbrv
\eqabbrvtrue    
\newcommand{\eqtext}{\ifeqabbrv Eq.\else Equation\fi}

\title{Evolving Collective Behavior in Self-Organizing Particle Systems}

\ifanon
\author{Anonymous Authors}
\else
\author{Devendra Parkar$^1$, Kirtus G.\ Leyba$^1$, Raylene A.\ Faerber$^1$, \and Joshua J.\ Daymude$^{1,*}$ \\
\mbox{} \\
$^1$ Biodesign Center for Biocomputing, Security and Society and School of Computing and Augmented Intelligence \\
Arizona State University, Tempe, AZ, United States \\
$^*$\texttt{jdaymude@asu.edu}}
\fi

\begin{document}

\maketitle

\begin{abstract}
    Local interactions drive emergent collective behavior, which pervades biological and social complex systems.
    But uncovering the interactions that produce a desired behavior remains a core challenge.
    In this paper, we present \evosops, an evolutionary framework that searches landscapes of stochastic distributed algorithms for those that achieve a mathematically specified target behavior.
    These algorithms govern \textit{self-organizing particle systems} (SOPS) comprising individuals with no persistent memory and strictly local sensing and movement.
    For \textit{aggregation}, \textit{phototaxing}, and \textit{separation} behaviors, \evosops\ discovers algorithms that achieve 4.2--15.3\% higher fitness than those from the existing ``stochastic approach to SOPS'' based on mathematical theory from statistical physics.
    \evosops\ is also flexibly applied to new behaviors such as \textit{object coating} where the stochastic approach would require bespoke, extensive analysis.
    Finally, we distill insights from the diverse, best-fitness genomes produced for aggregation across repeated \evosops\ runs to demonstrate how \evosops\ can bootstrap future theoretical investigations into SOPS algorithms for new behaviors.
\end{abstract}

\section{Introduction} \label{sec:intro}

When designing distributed systems---whether swarm robots, autonomous vehicular networks, or the Internet---self-organization in nature is a frequent inspiration~\citep{Gershenson2020-selforganizationartificial}.
From the beauty of starling murmurations~\citep{King2012-murmurations,Goodenough2017-birdsfeather} to the efficiency of ant colonies~\citep{Flanagan2012-quantifyingeffect,Cornejo2014-taskallocation}, sophisticated collective behaviors arise from individuals' local interactions, conferring inherent scalability and robustness on the system.
To engineer distributed systems that benefit from this paradigm, the core question is:
\textit{Given a desired collective behavior and any one individual's capabilities, what interactions among individuals are sufficient to yield the goal behavior?}

The focus of our research is using \textit{evolutionary search} to ``compile'' mathematically specified collective behaviors into interactions among mobile agents.
This goal is distinct from prior works that evolve decentralized controllers for robots that perform tasks individually~\citep{Scheper2016-behaviortrees,Phillips2021-evolutionaryrobotics}, modular robots with fixed morphologies~\citep{Weel2017-onlinegait,Kvalsund2022-centralizeddecentralized,Pigozzi2022-evolvingmodularity}, cellular automata that allow replication~\citep{Joachimczak2016-artificialmetamorphosis,Pena2021-lifeworth,Turney2021-measuringbehavioral,Khajehabdollahi2023-locallyadaptive}, and swarms that exhibit only one collective behavior~\citep{Hecker2015-pheromonesevolving,FernandezPerez2017-learningcollaborative,Hahn2020-foragingswarms,Mirhosseini2022-adaptivephototaxis}.
Inspired by the self-organizing capabilities of simple biological collectives (e.g., microbes, slime molds) and analog collectives without any explicit computational capabilities (e.g., grains of sand or other granular matter), we aim to make our interactions achievable by agents with strictly local sensing and movement, no persistent memory, and no explicit communication or identifiers.
Such computationally restricted agents cannot implement many of the recent successful evolutionary methodologies for swarms~\citep{Doursat2012-morphogeneticengineering}, including gradient computation~\citep{VanDiggelen2022-environmentinduced}, surprise minimization~\citep{Kaiser2022-innatemotivation}, and heterogeneous trait sharing~\citep{Hart2018-evolutionfunctionally,Wang2022-robotsmodels}.

We consider \textit{self-organizing particle systems} (SOPS), an abstraction of programmable matter in which computationally restricted particles move over a discrete lattice~\citep{Derakhshandeh2014-amoebotba,Daymude2019-computingprogrammable}.
Particles in a SOPS execute the same distributed algorithm that maps their current neighborhood configuration to a movement probability, yielding a memoryless, stochastic form of control (see $\S$\nameref{subsec:sops} for details).
Despite the particles' limitations, SOPS algorithms have been found for a variety of collective behaviors~\citep{Cannon2016-markovchain,AndresArroyo2018-stochasticapproach,Savoie2018-phototacticsupersmarticles,Cannon2019-localstochastic,Li2021-programmingactive,Kedia2022-localstochastic,Oh2023-adaptivecollective}.
These algorithms were developed using the \textit{stochastic approach to SOPS}~\citep{Cannon2016-markovchain} which uses mathematical theory from statistical physics to produce Markov chain algorithms with controlled long-run behaviors.
However, each application of this approach requires its own bespoke, extensive analysis.
Moreover, the stochastic approach can only produce a single algorithm per task, offering little insight into the diversity of interactions that can drive a collective behavior.

\paragraph{Our Contributions}

In this paper, we present \evosops, a high-performance evolutionary framework that searches landscapes of stochastic distributed SOPS algorithms for those that achieve a mathematically specified collective behavior ($\S$\nameref{subsec:framework}).
Our fitness functions reward algorithms that achieve a collective behavior \textit{scalably} (regardless of the number of particles) and \textit{robustly} (resilient to randomness in initial conditions and executions).
We apply \evosops\ to four collective behaviors: \textit{aggregation}, where particles gather compactly; \textit{phototaxing}, where particles simultaneously aggregate and move towards a light source; \textit{separation}, where heterogeneous particles aggregate as a whole and by color class; and \textit{object coating}, where particles form as many even layers as possible around a connected but arbitrarily shaped object ($\S$\nameref{subsec:behaviors}).
For the first three behaviors, \evosops\ discovers algorithms that achieve 4.2--15.3\% higher fitness than those developed using the theory-based stochastic approach to SOPS; it also finds the first known memoryless SOPS algorithms for object coating ($\S$\nameref{subsec:efficacy}).
Repeated \evosops\ runs for a given behavior explore \textit{diverse} regions of genome space, and the resulting exploration trajectories reveal differences in behavior complexity ($\S$\nameref{subsec:diversity}).
Finally, we analyze the best-fitness genomes for aggregation ($\S$\nameref{subsec:interpret}), revealing stark differences from the stochastic approach's algorithm and demonstrating how \evosops\ can be used to bootstrap future theoretical investigations into SOPS algorithms for new behaviors.

\section{Self-Organizing Particle Systems} \label{subsec:sops}

\begin{figure}[t]
    \centering
    \begin{subfigure}{0.48\columnwidth}
        \centering
        \includegraphics[height=3.25cm]{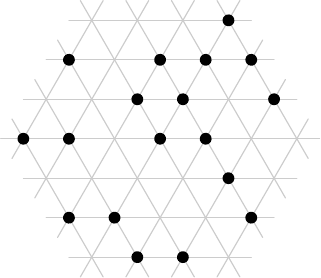}
        \caption{}
        \label{fig:sops:lattice}
    \end{subfigure}
    \hfill
    \begin{subfigure}{0.5\columnwidth}
        \centering
        \includegraphics[height=3.25cm]{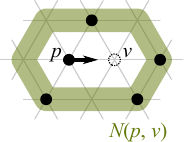}
        \caption{}
        \label{fig:sops:neighborhood}
    \end{subfigure}
    \caption{(a) A SOPS exists on the triangular lattice $\Gtri$, but (b) each particle $p$ can only view its extended neighborhood $N(p, v)$ to decide whether to move to an adjacent node $v$.}
    \label{fig:sops}
\end{figure}

A SOPS consists of individual, homogeneous computational elements called \textit{particles}~\citep{Derakhshandeh2014-amoebotba}.
Particles occupy nodes of the triangular lattice $\Gtri = (V, E)$, with at most one particle per node, and can move along its edges (\figtext~\ref{fig:sops:lattice}).
They have no persistent state, are anonymous (lacking unique identifiers), and do not have any global information such as a shared coordinate system or estimate of the total number of particles.

Given any particle $p$ and one of its six adjacent nodes $v$ in $\Gtri$, the \textit{extended neighborhood} $N(p, v) = N(p) \cup N(v) \setminus \{p, v\}$ is the set of nodes in $\Gtri$ adjacent to $p$ or $v$ (\figtext~\ref{fig:sops:neighborhood}).
The \textit{configuration} of an extended neighborhood describes the positions and types of particles it contains.
When clear from context, we may refer to extended neighborhoods and their configurations interchangeably.

We consider \textit{stochastic distributed algorithms} $\alg$ that map extended neighborhoods $N(p, v)$ to probabilities $\alg[N(p, v)] \in (0, 1]$ of particle $p$ moving to node $v$.
In a SOPS, every particle executes the same algorithm $\alg$ in a sequence of \textit{steps}.
At each step, a move is proposed by choosing a particle $p$ and one of its six adjacent nodes $v$ uniformly at random.
If this proposed move is \textit{valid}---a property that is behavior-specific---$p$ commits to the move with probability $\alg[N(p, v)]$.
Otherwise, it stays in its current position.

\section{Evolving SOPS Algorithms} \label{sec:evolvesops}

In this section, we describe our \evosops\ framework for evolving stochastic distributed algorithms for collective behavior ($\S$\nameref{subsec:framework}).
We then formally specify our four target behaviors: aggregation, phototaxing, separation, and object coating ($\S$\nameref{subsec:behaviors}).
We conclude with a brief description of the \evosops\ implementation, experiment environment, and performance metrics ($\S$\nameref{subsec:performance}).

\subsection{\texorpdfstring{The \evosops\ Framework}{The EvoSOPS Framework}} \label{subsec:framework}

\subsubsection{Genome Representation} \label{subsubsec:genome-representation}

\begin{figure}[t]
    \centering
    \includegraphics[width=\columnwidth]{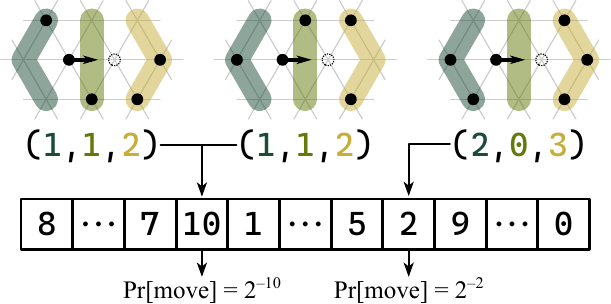}
    \caption{The genome representation of an algorithm $\alg$ is a list of integer-valued genes indexed by groups of extended neighborhoods.
    For example, the aggregation genome groups extended neighborhoods by the number of neighbors in their back (blue), middle (green), and front (yellow) regions.
    Loci vary by behavior.
    The alleles $i \in \{0, \ldots, 10\}$ correspond to movement probabilities $2^{-i} \in \alg$.}
    \label{fig:genome-representation}
\end{figure}

Recall that SOPS algorithms map extended neighborhood configurations to probabilities of movement ($\S$\nameref{subsec:sops}), suggesting a genome representation that uses a list of probabilities indexed by extended neighborhoods (\figtext~\ref{fig:genome-representation}).
To improve evolutionary search efficiency, we coarse-grain this large genome space in two ways.
First, we use integer-valued alleles $i \in \{0, \ldots, 10\}$ corresponding to movement probabilities $2^{-i}$.
Our experiments reveal that this geometric sequence of probabilities ranging from $2^0 = 1$ to $2^{-10} \approx 0.001$ sufficiently restricts the search space while still allowing for diverse solutions (see $\S$\nameref{sec:results}).
We omit a move probability of zero to avoid pathological deadlock configurations where no particle can move.

Second, we group functionally similar extended neighborhood configurations so they share the same move probability.
Specifically, we partition the eight nodes of an extended neighborhood $N(p, v)$ into three regions: \textit{back} (the three nodes adjacent only to $p$), \textit{middle} (the two nodes adjacent to both $p$ and $v$), and \textit{front} (the three nodes adjacent only to $v$).
Tuples of behavior-specific back/middle/front information serve as loci.
For example, aggregation loci are triples counting the neighbors in each region, ignoring those neighbors' exact positions (\figtext~\ref{fig:genome-representation}).
This reduces the length of aggregation genomes from $2^8 = 256$ distinct extended neighborhood configurations to $48$ loci.
Details of each behavior's loci are given in $\S$\nameref{subsec:behaviors}.

\subsubsection{Fitness Evaluation} \label{subsubsec:fitness}

To evaluate a genome, we execute the corresponding stochastic distributed algorithm in a SOPS.
Any high-fitness algorithm should be both \textit{robust} to the system's inherent stochasticity and \textit{scalable}, performing well across SOPS systems of different numbers of particles.
For those reasons, our fitness evaluation considers three particle system sizes $\sopssizes$ and, for each system size $\numparticles \in \sopssizes$, executes $\numtrials = 3$ independent trials of the algorithm.\footnote{The exact number of particles varies by behavior: aggregation and phototaxing use $S = \{61, 169, 271\}$ which aggregate ideally as regular hexagons, separation uses one less particle per size so the $\numcolors = 3$ color classes are evenly sized, and object coating uses $S = \{66, 144, 252\}$ to make room for the object.
See $\S$\nameref{subsec:behaviors} for details.}

Each trial randomly initializes a SOPS system within a regular hexagonal arena.
For consistency across system sizes $\numparticles$, the arena size is set such that the density of particles in the arena is roughly $\density = 0.5$.
The $\numparticles$ particles are then placed at unoccupied positions chosen uniformly at random within the arena.
The algorithm $\alg$ is then executed as described in $\S$\nameref{subsec:sops}: at each step, a particle $p$ and an adjacent node $v$ are chosen uniformly at random and $p$ moves to $v$ if the move is valid with the corresponding probability $\alg[N(p, v)]$.
All algorithm executions run for $\numparticles^3$ steps, as this is asymptotically comparable to the runtime of the algorithms produced by the stochastic approach~\citep{Cannon2016-markovchain,Cannon2019-localstochastic,Li2021-programmingactive}.

Each behavior $b$ defines a \textit{quality measure} $q_b(\sigma)$ quantifying how well a SOPS configuration $\sigma$ achieves behavior $b$.
For example, aggregation's quality measure $q_\text{agg}(\sigma)$ counts the number of neighboring particle pairs in $\sigma$; this rewards higher values to configurations that are better aggregated.
Multi-objective behaviors---such as phototaxing which both aggregates and moves towards light---define quality measures $q_{b,i}$ and corresponding weights $w_i \in (0, 1)$ for each of their $K$ objectives, where $\sum_{i=1}^K w_i = 1$.
Let $\sigma_{\numparticles,j}$ be the final configuration of the $j$-th trial on $\numparticles$ particles and let $\sigma_b^*(\numparticles)$ be any configuration of $\numparticles$ particles maximizing the quality measure(s).
A genome's fitness value is calculated as an average of quality ratios over all trials and system sizes, which has an optimal value of 1: 
\begin{equation} \label{eq:fitness}
    F_b = \frac{1}{|\sopssizes| \cdot \numtrials} \sum_{\numparticles \in \sopssizes} \sum_{j=1}^{\numtrials} \sum_{i=1}^K w_i \cdot \frac{q_{b,i}(\sigma_{\numparticles,j})}{q_{b,i}\left(\sigma_b^*(\numparticles)\right)}
\end{equation}

\subsubsection{Selection} \label{subsubsec:selection}

\evosops\ uses deterministic tournament selection with tournament size two to determine which algorithm genomes move forward with crossover and mutation.
In each tournament, a pair of genomes is chosen uniformly at random from the population and the one with higher fitness is selected.
We opted for this selection operator to give lower-fitness genomes an opportunity to improve, increasing exploration of the algorithm landscape.

\subsubsection{Crossover} \label{subsubsec:crossover}

\evosops\ uses standard two-point crossover.
Our genomes are 1D lists of integers (corresponding to probabilities), so we simply choose two points uniformly at random and exchange the resulting partitions.
We use two-point crossover because the lexicographical ordering of loci puts genes for similar extended neighborhoods nearby each other.

\subsubsection{Mutation} \label{subsubsec:mutation}

A mutation to an allele $a \in \{0, \ldots, 10\}$ updates
\begin{equation} \label{eq:mutation}
    a \gets \left\{ \begin{array}{cl}
        \max\{a - 1, 0\} & \text{with probability } 1/2; \\
        \min\{a + 1, 10\} & \text{otherwise.}
    \end{array} \right.
\end{equation}
Each gene mutates independently with a behavior-specific probability $\mutrate \in (0, 1)$.
For aggregation and phototaxing, we target one mutation per genome per generation; for separation and object coating---which have much larger genomes---we target three (see Table~\ref{tab:params}).

\evosops\ also includes hypermutation to discourage premature convergence~\citep{Srinivas1994-adaptiveprobabilities}.
It is based on \textit{population diversity}, defined as the average pairwise $L_1$-distance among the population's genomes normalized by the maximum possible $L_1$-distance between two genomes in the population.
So population diversity ranges in $[0, 1]$, where $0$ indicates a population of identical genomes.
When a hypermutation factor $\hypermut > 1$ is given, \evosops\ changes the mutation rate to $\hypermut \cdot \mutrate$ if population diversity drops to a specified lower threshold and resets it to $\mutrate$ if the population diversity recovers to a specified upper threshold.

\subsection{Specifying Collective Behaviors} \label{subsec:behaviors}

With the \evosops\ framework in place, we turn to the behavior-specific details of aggregation, phototaxing, separation, and object coating.
Table~\ref{tab:params} summarizes all \evosops\ parameter values across these four behaviors.

\begin{table}[t]
    \centering
    \caption{\evosops\ parameters for each collective behavior.}
    \label{tab:params}
    \begin{tabular}{lccccc}
        \toprule
        \textbf{Parameter} & \textbf{Agg.} & \textbf{Ptx.} & \textbf{Sep.} & \textbf{Coat.} \\
        \midrule
        genome length & 48 & 144 & 600 & 600 \\
        population size ($\population$) & 50 & 150 & 600 & 600 \\
        \# generations ($\numgens$) & 100 & 200 & 300 & 750 \\
        mutation rate ($\mutrate$) & 0.021 & 0.007 & 0.005 & 0.005 \\
        hypermutation ($\hypermut$) & --- & --- & 10x & 10x \\
        \bottomrule
    \end{tabular}
\end{table}

\subsubsection{Aggregation} \label{subsubsec:aggregation}

\begin{figure}[t]
    \centering
    \begin{subfigure}{0.48\columnwidth}
        \centering
        \includegraphics[width=0.6\textwidth]{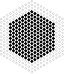}
        \caption{$\sigma_{\text{agg}}^*(\numparticles = 169)$}
        \label{fig:optconfig:agg}
    \end{subfigure}
    \hfill
    \begin{subfigure}{0.48\columnwidth}
        \centering
        \includegraphics[width=0.6\textwidth]{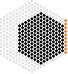}
        \caption{$\tilde{\sigma}_{\text{ptx}}^*(\numparticles = 169)$}
        \label{fig:optconfig:ptx}
    \end{subfigure}
    \\ \bigskip
    \begin{subfigure}{0.48\columnwidth}
        \centering
        \includegraphics[width=0.6\textwidth]{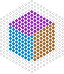}
        \caption{$\tilde{\sigma}_{\text{sep}}^*(\numparticles = 168, \numcolors = 3)$}
        \label{fig:optconfig:sep}
    \end{subfigure}
    \hfill
    \begin{subfigure}{0.48\columnwidth}
        \centering
        \includegraphics[width=0.6\textwidth]{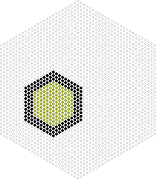}
        \caption{$\sigma_{\text{coat}}^*(\numparticles = 144)$}
        \label{fig:optconfig:coat}
    \end{subfigure}
    \caption{The (near-)optimal configurations used to normalize the quality measures for (a) aggregation, (b) phototaxing, (c) separation, and (d) object coating.}
    \label{fig:optconfig}
\end{figure}

To aggregate, particles gather compactly anywhere within the arena, akin to how schools of fish swarm to intimidate predators~\citep{Magurran1990-adaptivesignificance} or how ants form rafts to survive floods~\citep{Mlot2011-fireants}.
Following~\citet{Li2021-programmingactive}, we define aggregation as the maximization of $q_\text{agg}(\sigma) = e(\sigma)$, the number of lattice edges in $\sigma$ with both endpoints occupied by particles.
The optimally aggregated configuration $\sigma_\text{agg}^*(\numparticles)$ on $\numparticles$ particles is a regular hexagon, or as close to one as possible w.r.t.\ $\numparticles$ (\figtext~\ref{fig:optconfig:agg}).
Substituting $q_\text{agg}$ and $\sigma_\text{agg}^*(\numparticles)$ into \eqtext~\ref{eq:fitness} yields $F_\text{agg}$, the aggregation fitness function.
The aggregation genome is indexed by triples indicating the number of neighbors a particle has in its back, middle, and front extended neighborhood regions (\figtext~\ref{fig:genome-representation}), totaling 48 genes.
A move into a node $v$ is considered valid if and only if $v$ is unoccupied.

\subsubsection{Phototaxing} \label{subsubsec:phototaxing}

In phototaxing, a SOPS must simultaneously aggregate and move towards a set of light sources $L \subseteq V$ along one wall of the hexagonal arena.
We model this multi-objective optimization as a pair of quality measures: $q_\text{ptx,1} = q_\text{agg}$ for aggregation ($w_1 = 0.75$) and $q_\text{ptx,2}$ for directed motion ($w_2 = 0.25)$.\footnote{Preliminary experiments indicated that a moderate bias towards aggregation over the second objective (directed motion for phototaxing and chromatic sorting for separation) best aligned fitness with the ideal configurations $\tilde{\sigma}_\text{ptx}^*(n)$ and $\tilde{\sigma}_\text{sep}^*(n, c)$.}
For any node $v \not\in L$, let $\phi_\ell(v)$ be the ``height'' of $v$, where nodes adjacent to the wall opposite $L$ have height 1, the next row of nodes have height 2, and so on until the nodes adjacent to $L$ which have maximum height.
We define $q_\text{ptx,2}(\sigma) = (1/\numparticles) \sum_{p \in \sigma} \phi_\ell(p)$ as the average particle height.
The optimal configuration $\sigma_\text{ptx}^*(\numparticles)$ is not so easily defined, as it depends on the exact number of particles $\numparticles$ and the choice of weights $w_i$.
We instead use a reasonable approximation, $\tilde{\sigma}_\text{ptx}^*(\numparticles)$, which places a (near-)regular hexagon of $\numparticles$ particles adjacent to $L$ (\figtext~\ref{fig:optconfig:ptx}).
Substituting these $q_\text{ptx}$ and $\tilde{\sigma}_\text{ptx}^*(\numparticles)$ into \eqtext~\ref{eq:fitness} yields $F_\text{ptx}$, the phototaxing fitness function.

The genome for phototaxing generalizes the one for aggregation by allowing particles to sense light.
Each light source emits a ray of light extending perpendicular from its wall.
Particles occlude one another, so only the closest particle along any one ray of light senses it, and any others behind it do not.
Thus, in addition to a particle's number of back, middle, and front neighbors, loci tuples for phototaxing include a fourth entry indicating whether the particle is moving from a lit position to an unlit position, an unlit position to a lit position, or between positions with the same light status.
This totals 144 genes.
As in aggregation, a move into a node $v$ is considered valid if and only if $v$ is unoccupied.

\subsubsection{Separation} \label{subsubsec:separation}

In separation, each particle has one of $\numcolors > 1$ colors and must aggregate as a whole and by color.
In nature, separation occurs at the boundary of inter- and intra-group dynamics, such as when bacteria compete for resources within biofilms~\citep{Stewart2008-physiologicalheterogeneity,Wei2015-moleculartweeting} or when social insects engage in conflict with other colonies~\citep{Roulston2003-nestmatediscrimination,Johnson2011-nestmaterecognition}.
Following~\citet{Cannon2019-localstochastic}, we define separation in SOPS as the simultaneous maximization of $q_\text{sep,1}(\sigma) = e(\sigma)$---the number of neighboring particle pairs---and $q_\text{sep,2}(\sigma) = h(\sigma)$, the number of lattice edges in $\sigma$ whose endpoints are occupied by particles of the same color.
We fix weights $w_1 = 0.65$ and $w_2 = 0.35$.\footnotemark[3]
As in phototaxing, it is difficult to define the optimally separated configuration $\sigma_\text{sep}^*(\numparticles, \numcolors)$ for $\numparticles$ particles and $\numcolors$ colors since it depends on $\numparticles$, $\numcolors$, and $w_i$.
Our approximation $\tilde{\sigma}_\text{sep}^*(\numparticles, \numcolors)$ is the (near-)regular hexagon of $\numparticles$ particles tiled by $\numcolors$ convex monochromatic regions (\figtext~\ref{fig:optconfig:sep}).
Substituting these $q_\text{sep}$ and $\tilde{\sigma}_\text{sep}^*(\numparticles, \numcolors)$ into \eqtext~\ref{eq:fitness} yields $F_\text{sep}$, the separation fitness function.

Genomes for separation algorithms extend the basic idea of the aggregation genome by allowing particles to differentiate among neighbors by color. 
Specifically, loci tuples contain a particle's number of neighbors and number of same-color neighbors for each of its back, middle, and front regions, totaling 600 genes.
Due to this large space of genomes, separation uses a hypermutation factor of $H = 10 \times$ triggered at a lower threshold of $0.072$ population diversity and reset at an upper threshold of $0.27$.

Again following~\citet{Cannon2019-localstochastic}, a particle $p$ of color $i$ moving into a node $v$ is valid if either $v$ is unoccupied or $v$ is occupied by a particle $p'$ of color $j \neq i$.
The latter is called a \textit{swap move} which exchanges the positions of $p$ and $p'$.
Swap moves are uninteresting between same-color neighbors, but enable heterogeneous neighbors to make progress even when surrounded.
\evosops\ enacts swap moves with probability $\min\{\alg[N(p, p')], \alg[N(p', p)]\}$.

\subsubsection{Object Coating} \label{subsubsec:coating}

In object coating, the SOPS must form as many even layers as possible around a connected object $O \subseteq V$ in the arena.
Formally, for any node $v \not\in O$, let $d_o(v)$ be the shortest distance (in edges) from $v$ to any node in $O$.
We define $q_\text{coat}(\sigma) = 1 / \sum_{p \in \sigma} d_o(p)$ as the reciprocal of the sum over all particle-to-object distances; the reciprocal simply turns what would naturally be a problem of minimizing distances into a corresponding fitness maximization problem.
Regardless of the object's shape and size, the optimal configuration $\sigma_\text{coat}^*(\numparticles)$ on $\numparticles$ particles places particles on the nodes $v \not\in O$ with the $\numparticles$ smallest $d_o(v)$ values (\figtext~\ref{fig:optconfig:coat}).
Substituting these $q_\text{coat}$ and $\sigma_\text{coat}^*(\numparticles)$ into \eqtext~\ref{eq:fitness} yields $F_\text{coat}$, the coating fitness function.

The object coating genome has analogous structure to the separation genome, differentiating between particles and objects in the extended neighborhood.
Loci contain a particle's numbers of particle neighbors and object neighbors in each of its back, middle, and front regions, totaling 600 genes.
A move into a node $v$ is considered valid if and only if $v$ is unoccupied (by both particles and objects).

\subsection{Implementation and Performance} \label{subsec:performance}

\begin{table}[t]
    \centering
    \caption{Execution times for a fitness evaluation of $P = 1$ vs.\ $P = 128$ individuals (in seconds), scaling factors with respect to ideal parallel scaling, and parallel runtime as a \% of the corresponding sequential runtime.}
    \label{tab:performance}
    \begin{tabular}{lcccc}
        \toprule
        \textbf{Behavior} & $P = 1$ & $P = 128$ & \textbf{Scaling} & \textbf{\% Seq.} \\
        \midrule
        Aggregation & 8 s & 73 s & 4.56x & 7.13\% \\
        Phototaxing & 25 s & 208 s & 4.16x & 6.50\% \\
        Separation & 24 s & 239 s & 4.98x & 7.78\% \\
        Coating & 12 s & 106 s & 4.42x & 6.90\% \\
        \bottomrule
    \end{tabular}
\end{table}

\evosops\ is implemented in Rust to exploit the parallelization of population fitness evaluations; gain the efficiency of strongly typed, low-level languages such as C and C++; and guarantee memory safety at compile-time.
With each individual fitness evaluation requiring multiple SOPS execution trials for tens of millions of steps---and having to repeat this for hundreds of individuals per population across hundreds of generations---efficiency is paramount.
Table~\ref{tab:performance} reports the execution time for evaluating the fitness of a single individual vs.\ a population of 128 individuals for each of the four behaviors using 64 threads on a machine with one 64-core AMD EPYC 7713 processor and 50 GiB of memory.
The ideal runtime scaling for 128 fitness evaluations over 64 threads is twice that of a single evaluation; normalizing for this factor of two, we observe scaling factors of 4.16--4.98x the ideal.
Moreover, on this machine, \evosops\ cuts the execution time of 128 fitness evaluations to just 6.5--7.78\% of what would otherwise be required by a sequential execution of the same number of evaluations.
These optimizations brought a single complete \evosops\ run for separation (the most compute-intensive behavior) to roughly four days; thus, all subsequent results are reported for three independent \evosops\ runs per behavior.

\section{Results} \label{sec:results}

In $\S$\nameref{subsec:efficacy}, we demonstrate that \evosops\ discovers high-fitness algorithms for all four collective behaviors that outperform those from the stochastic approach to SOPS (where applicable) and retain their quality even when scaled to much larger SOPS sizes than their fitness evaluations considered.
In $\S$\nameref{subsec:diversity}, we characterize \evosops's explorations of genome space show that repeated runs discover diverse solutions to the same collective behavior. 
Finally, in $\S$\nameref{subsec:interpret}, we interpret high-fitness genomes for aggregation
as understandable insights.

\subsection{Efficacy and Scalability} \label{subsec:efficacy}

\begin{figure}[t!]
    \centering
    \begin{tikzpicture}
        \node[rotate=90] at (-4, 0) {\small \textbf{(a) Aggregation}};
        \node at (0, 0) {\includegraphics[height=4cm]{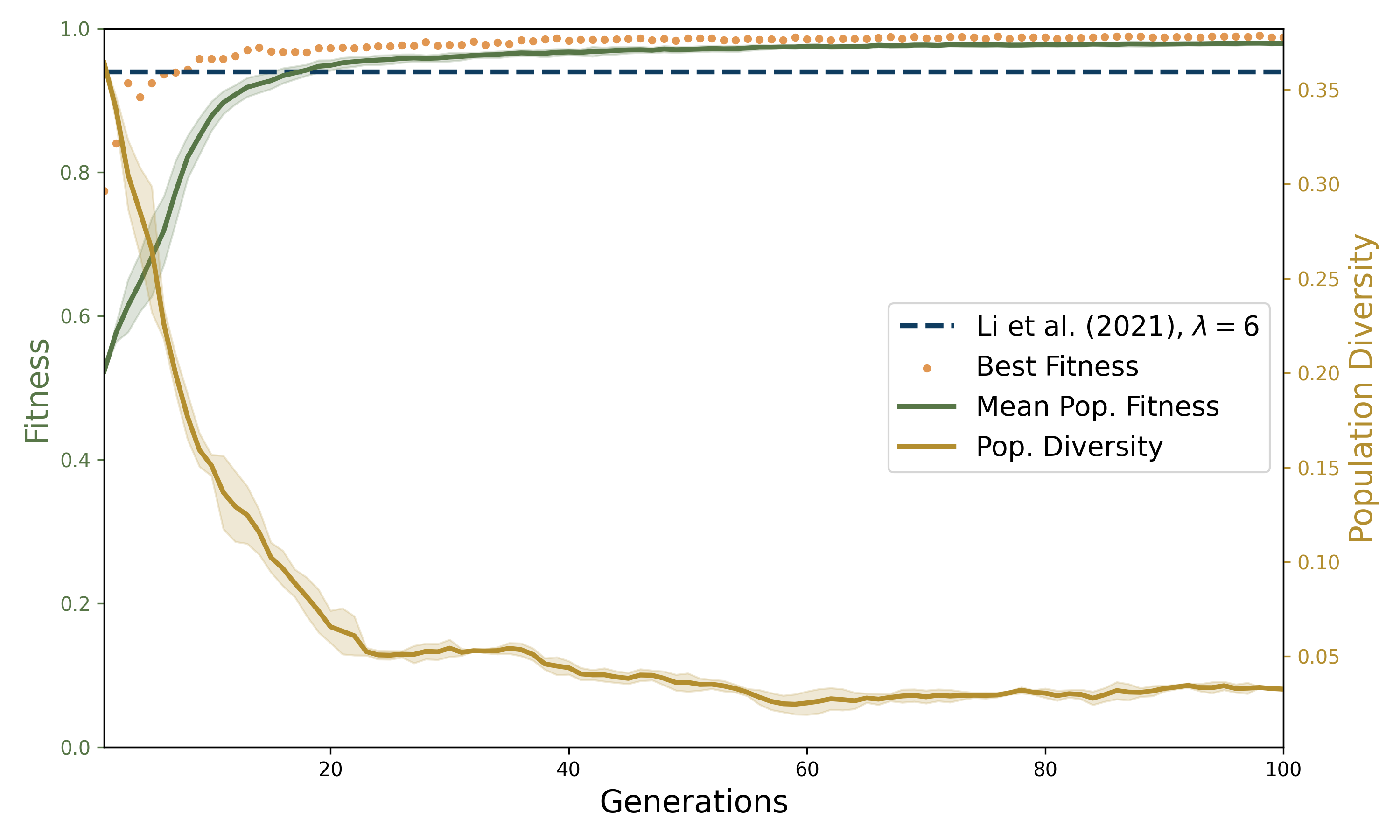}};

        \node[rotate=90] at (-4, -4) {\small \textbf{(b) Phototaxing}};
        \node at (0, -4) {\includegraphics[height=4cm]{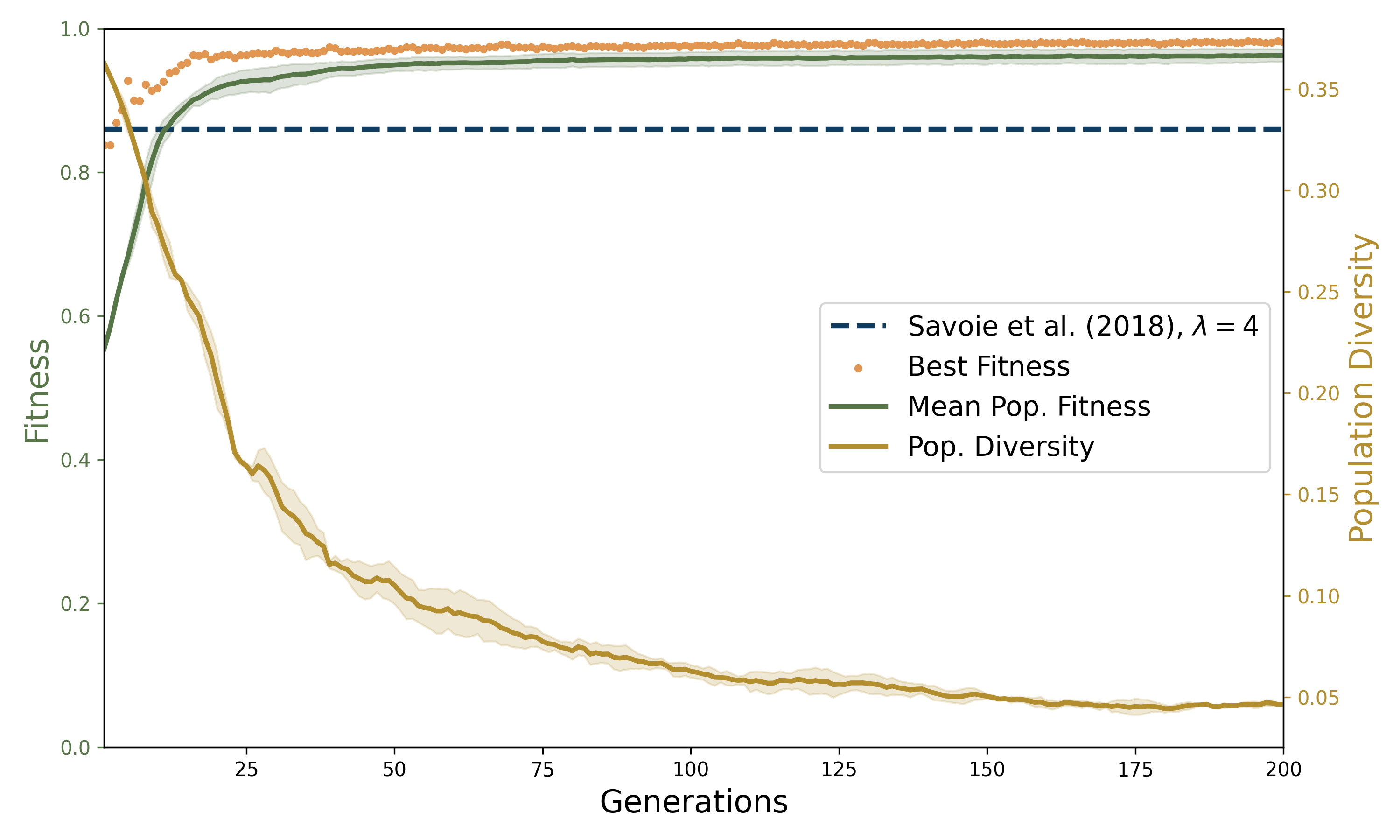}};

        \node[rotate=90] at (-4, -8) {\small \textbf{(c) Separation}};
        \node at (0, -8) {\includegraphics[height=4cm]{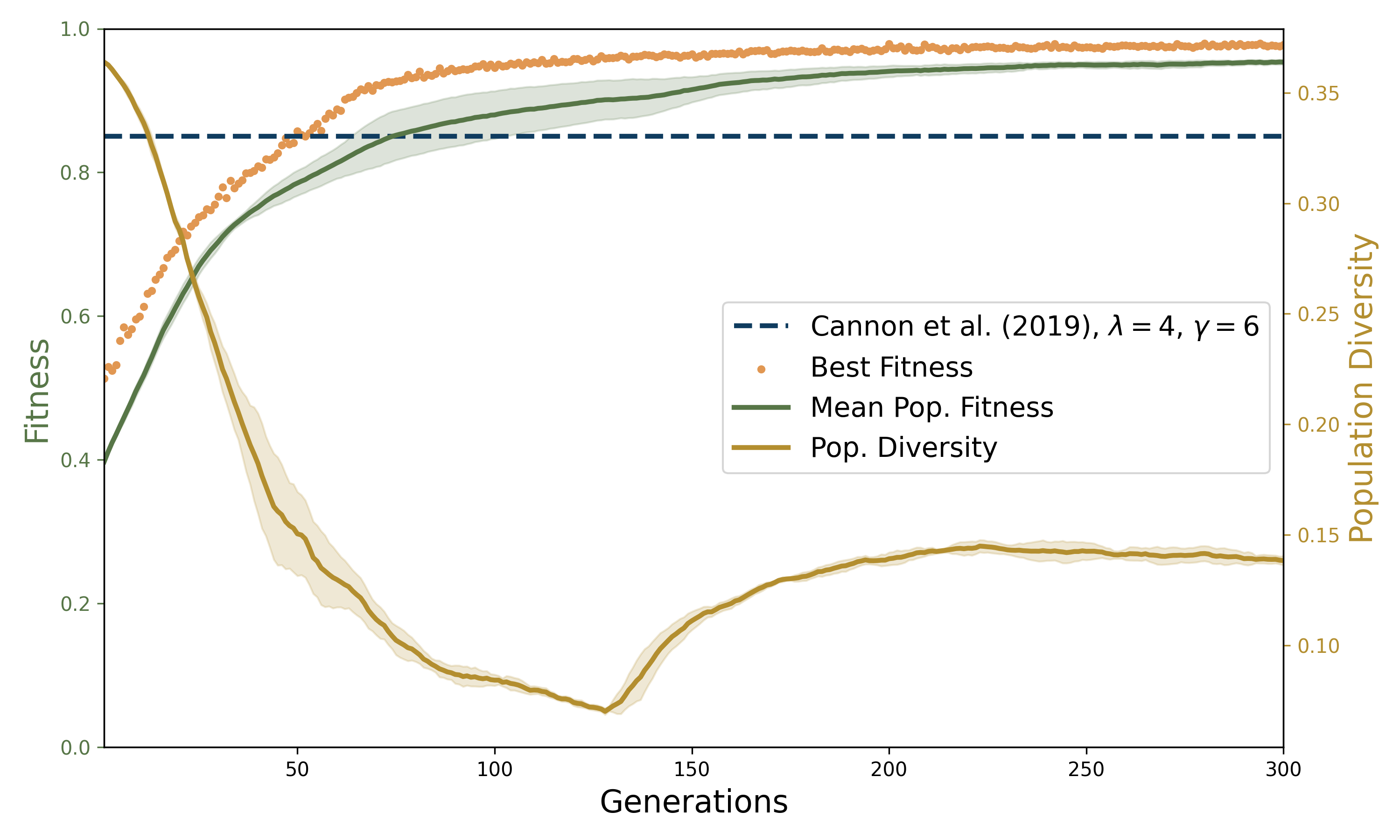}};

        \node[rotate=90] at (-4, -12) {\small \textbf{(d) Object Coating}};
        \node at (0, -12) {\includegraphics[height=4cm]{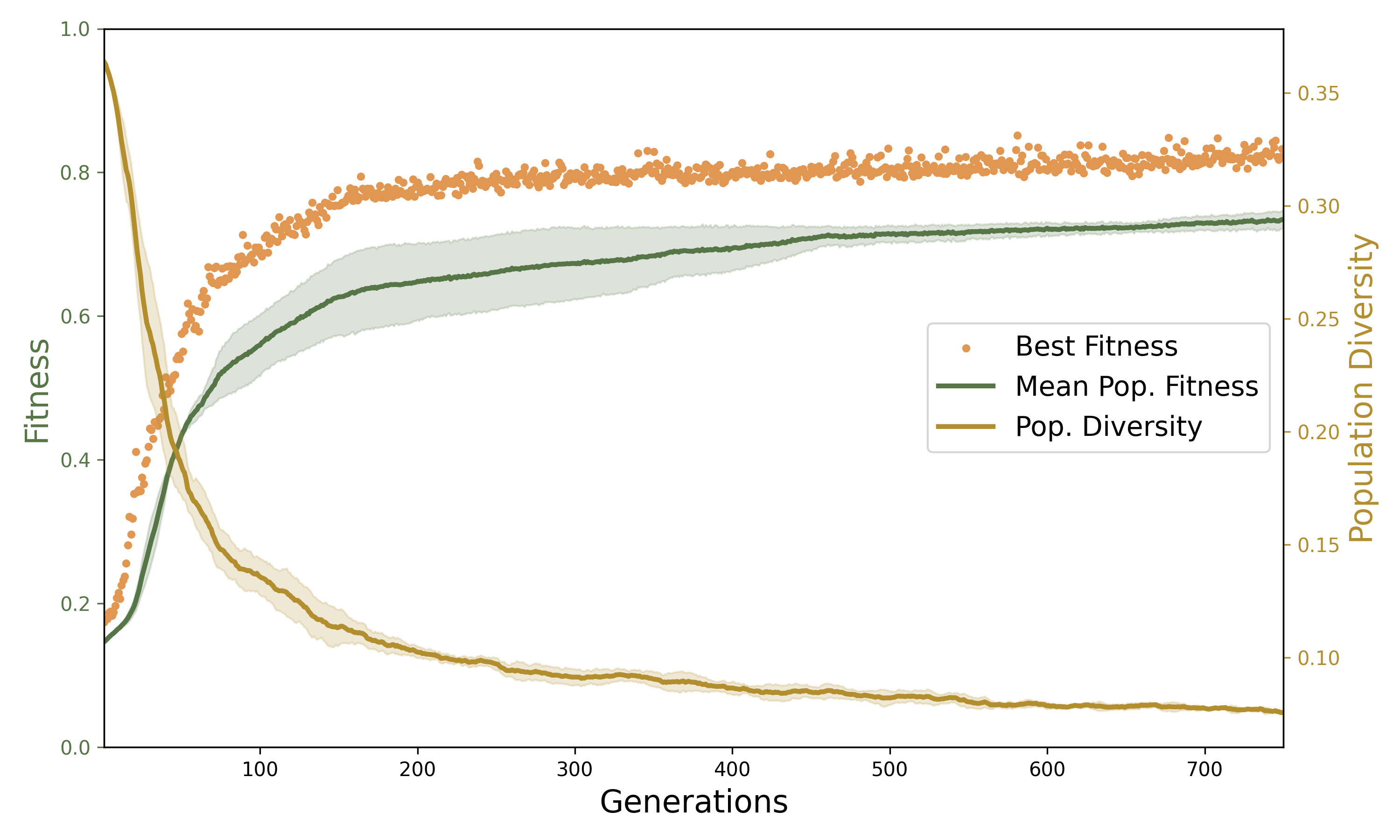}};
    \end{tikzpicture}
    \vspace{-0.4cm}
    \caption{Best fitness (orange dots), population fitness (mean $\pm$ std.\ dev., green), and population diversity (mean $\pm$ std.\ dev., gold) progressions for three \evosops\ runs per collective behavior.
    Fitness values for the stochastic approach's aggregation~\citep{Li2021-programmingactive}, phototaxing~\citep{Savoie2018-phototacticsupersmarticles}, and separation~\citep{Cannon2019-localstochastic} algorithms are shown as dark blue dashed lines.}
    \label{fig:fitnessdiversity}
\end{figure}
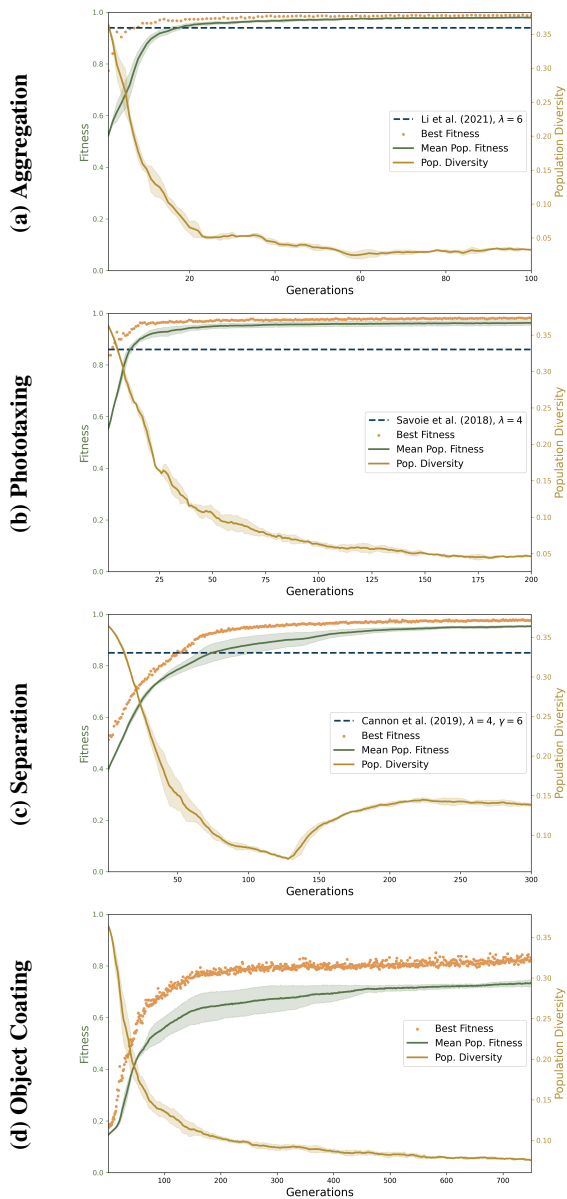

\begin{figure}[t]
    \centering
    \includegraphics[width=0.9\columnwidth]{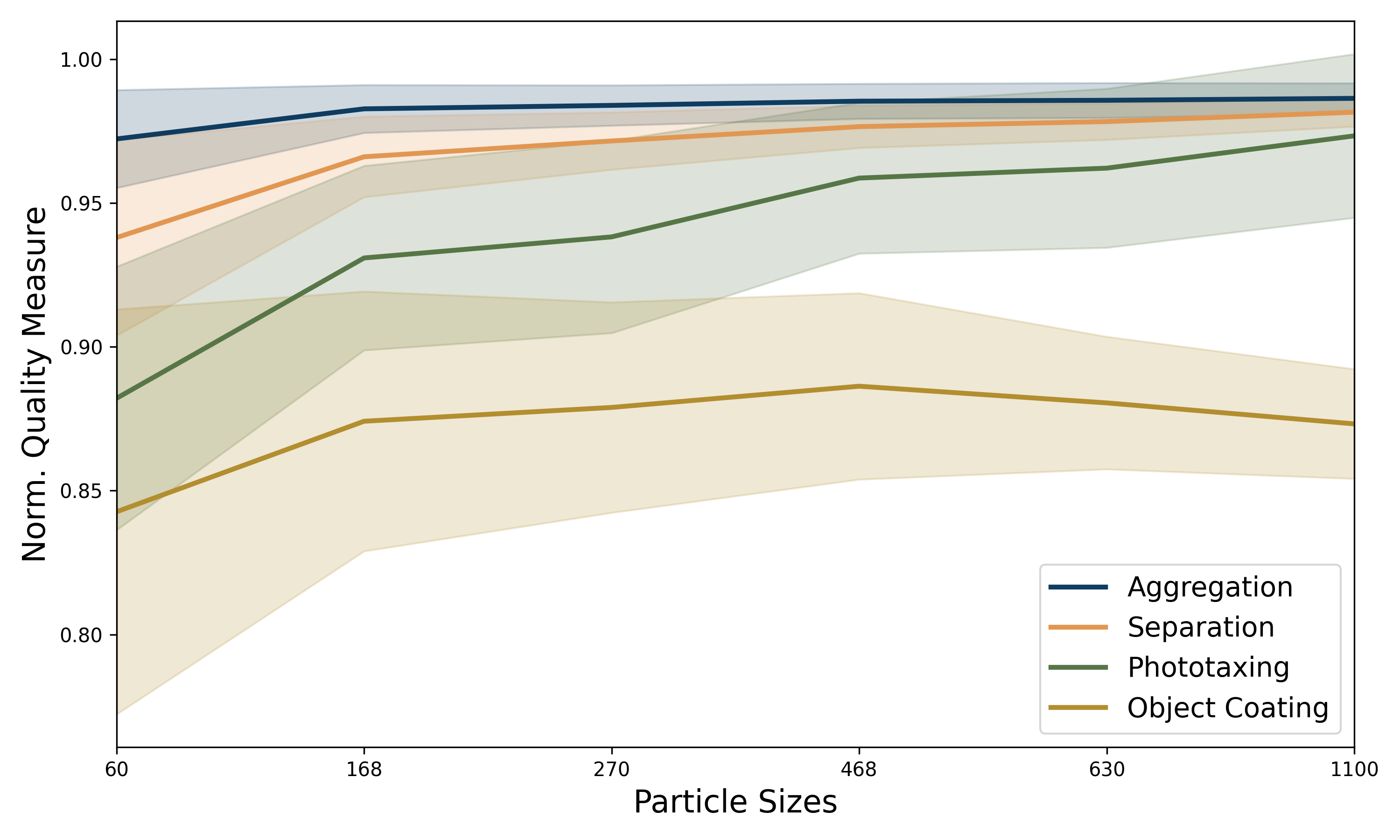}
    \vspace{-0.3cm}
    \caption{Mean $\pm$ standard deviation of normalized quality measures $\sum_{i=1}^K w_i \cdot \frac{q_{b,i}(\sigma(\numparticles))}{q_{b,i}(\sigma_b^*(\numparticles))}$ of 100 SOPS executions of each of the top five fitness genomes per \evosops\ run for each collective behavior $b$ over a range of SOPS sizes $\numparticles$.}
    \label{fig:scaleruns}
\end{figure}

\begin{table}[t]
    \centering
    \caption{Fitness values for the best-fitness \evosops\ genomes, mean fitness across all genomes in the final populations, fitness values attained by the theoretical algorithms designed using the stochastic approach to SOPS (where applicable), and fitness improvements of the best \evosops\ genomes over the theoretical algorithms.}
    \label{tab:summary}
    \begin{tabular}{lcccc}
        \toprule
        \textbf{Behavior} & \textbf{Best} & \textbf{Final} & \textbf{Theory} & \textbf{\% Improve} \\
        \midrule
        Aggregation & 0.99 & 0.98 & 0.95 & 4.21\% \\
        Phototaxing & 0.98 & 0.96 & 0.86 & 13.95\% \\
        Separation & 0.98 & 0.95 & 0.85 & 15.29\% \\
        Coating & 0.85 & 0.73 & --- & --- \\
        \bottomrule
    \end{tabular}
\end{table}

\begin{figure*}[t]
    \centering
    \begin{subfigure}{0.25\textwidth}
        \centering
        \begin{tikzpicture}
            \node[rotate=90] at (-2,0) {\small \textbf{Aggregation}};
            \node at (0,0) {\includegraphics[height=2.75cm]{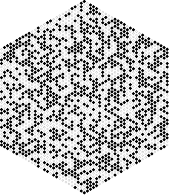}};
        \end{tikzpicture}
        \vspace{-0.15cm}
        \caption{Initialization, $F_\text{agg} = 0.5$}
        \label{fig:bigconfigs:agg:0}
    \end{subfigure}
    \hfill
    \begin{subfigure}{0.22\textwidth}
        \centering
        \includegraphics[height=2.75cm]{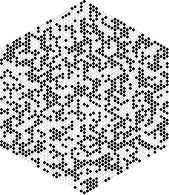}
        \caption{$\numparticles$ steps, $F_\text{agg} = 0.54$}
        \label{fig:bigconfigs:agg:n}
    \end{subfigure}
    \hfill
    \begin{subfigure}{0.22\textwidth}
        \centering
        \includegraphics[height=2.75cm]{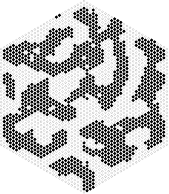}
        \caption{$\numparticles^2$ steps, $F_\text{agg} = 0.86$}
        \label{fig:bigconfigs:agg:n2}
    \end{subfigure}
    \hfill
    \begin{subfigure}{0.22\textwidth}
        \centering
        \includegraphics[height=2.75cm]{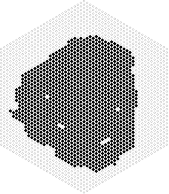}
        \caption{$\numparticles^3$ steps, $F_\text{agg} = 0.99$}
        \label{fig:bigconfigs:agg:n3}
    \end{subfigure}
    \\ \medskip
    \begin{subfigure}{0.25\textwidth}
        \centering
        \begin{tikzpicture}
            \node[rotate=90] at (-2,0) {\small \textbf{Phototaxing}};
            \node at (0,0) {\includegraphics[height=2.75cm]{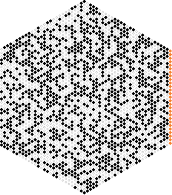}};
        \end{tikzpicture}
        \vspace{-0.15cm}
        \caption{Initialization, $F_\text{ptx} = 0.62$}
        \label{fig:bigconfigs:ptx:0}
    \end{subfigure}
    \hfill
    \begin{subfigure}{0.22\textwidth}
        \centering
        \includegraphics[height=2.75cm]{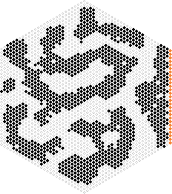}
        \caption{$\numparticles^2$ steps, $F_\text{ptx} = 0.79$}
        \label{fig:bigconfigs:ptx:n2}
    \end{subfigure}
    \hfill
    \begin{subfigure}{0.22\textwidth}
        \centering
        \includegraphics[height=2.75cm]{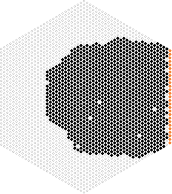}
        \caption{$20\numparticles^2$ steps, $F_\text{ptx} = 0.95$}
        \label{fig:bigconfigs:ptx:2n2}
    \end{subfigure}
    \hfill
    \begin{subfigure}{0.22\textwidth}
        \centering
        \includegraphics[height=2.75cm]{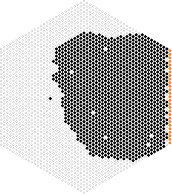}
        \caption{$\numparticles^3$ steps, $F_\text{ptx} = 0.97$}
        \label{fig:bigconfigs:ptx:n3}
    \end{subfigure}
    \\ \medskip
    \begin{subfigure}{0.25\textwidth}
        \centering
        \begin{tikzpicture}
            \node[rotate=90] at (-2,0) {\small \textbf{Separation}};
            \node at (0,0) {\includegraphics[height=2.75cm]{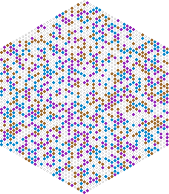}};
        \end{tikzpicture}
        \vspace{-0.15cm}
        \caption{Initialization, $F_\text{sep} = 0.39$}
        \label{fig:bigconfigs:sep:0}
    \end{subfigure}
    \hfill
    \begin{subfigure}{0.22\textwidth}
        \centering
        \includegraphics[height=2.75cm]{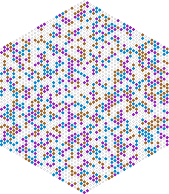}
        \caption{$\numparticles$ steps, $F_\text{sep} = 0.43$}
        \label{fig:bigconfigs:sep:n}
    \end{subfigure}
    \hfill
    \begin{subfigure}{0.22\textwidth}
        \centering
        \includegraphics[height=2.75cm]{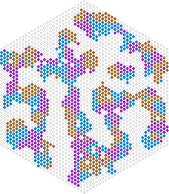}
        \caption{$\numparticles^2$ steps, $F_\text{sep} = 0.77$}
        \label{fig:bigconfigs:sep:n2}
    \end{subfigure}
    \hfill
    \begin{subfigure}{0.22\textwidth}
        \centering
        \includegraphics[height=2.75cm]{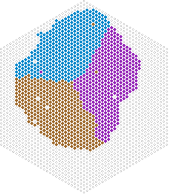}
        \caption{$\numparticles^3$ steps, $F_\text{sep} = 0.98$}
        \label{fig:bigconfigs:sep:n3}
    \end{subfigure}
    \\ \medskip
    \begin{subfigure}{0.25\textwidth}
        \centering
        \begin{tikzpicture}
            \node[rotate=90] at (-2,0) {\small \textbf{Object Coating}};
            \node at (0,0) {\includegraphics[height=2.75cm]{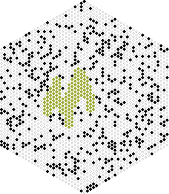}};
        \end{tikzpicture}
        \vspace{-0.15cm}
        \caption{Initialization, $F_\text{coat} = 0.31$}
        \label{fig:bigconfigs:coat:0}
    \end{subfigure}
    \hfill
    \begin{subfigure}{0.22\textwidth}
        \centering
        \includegraphics[height=2.75cm]{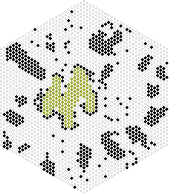}
        \caption{$\numparticles^2$ steps, $F_\text{coat} = 0.32$}
        \label{fig:bigconfigs:coat:n}
    \end{subfigure}
    \hfill
    \begin{subfigure}{0.22\textwidth}
        \centering
        \includegraphics[height=2.75cm]{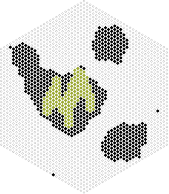}
        \caption{$60\numparticles^2$ steps, $F_\text{coat} = 0.32$}
        \label{fig:bigconfigs:coat:n2}
    \end{subfigure}
    \hfill
    \begin{subfigure}{0.22\textwidth}
        \centering
        \includegraphics[height=2.75cm]{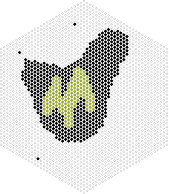}
        \caption{$\numparticles^3$ steps, $F_\text{coat} = 0.71$}
        \label{fig:bigconfigs:coat:n3}
    \end{subfigure}
    \caption{Time evolutions of a simulated SOPS executing the best-fitness \evosops\ algorithm for (a)--(d) aggregation on $\numparticles = \numprint{1141}$ particles, (e)--(h) phototaxing on $\numparticles = \numprint{1141}$ particles with the light sources (orange) on the right, (i)--(l) separation on $\numparticles = \numprint{1140}$ particles and $\numcolors = 3$ colors, and (m)--(p) object coating on $\numparticles = \numprint{500}$ particles and an irregular object (green).}
    \label{fig:bigconfigs}
\end{figure*}

\begin{figure}[t!]
    \centering
    \ifanon
    \includegraphics[angle=90,origin=c,width=0.7\columnwidth]{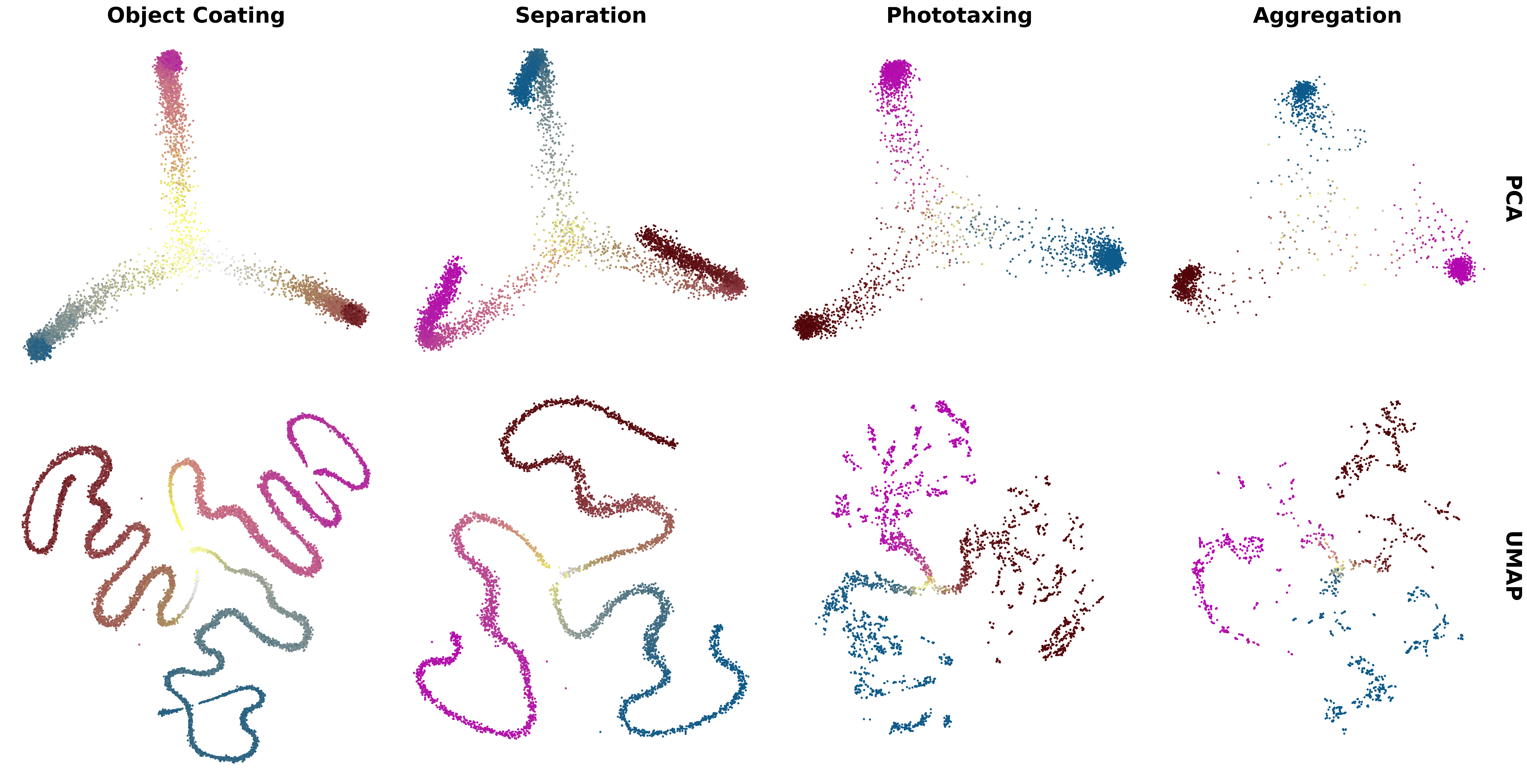}
    \else
    \includegraphics[angle=90,origin=c,width=0.9\columnwidth]{all_all.png}
    \fi
    \caption{PCA (left) and UMAP (right) embeddings of all generations' genomes for three independent \evosops\ runs per collective behavior.
    Each run is shown as a distinct color map, with higher color intensity indicating higher fitness.}
    \label{fig:diversity}
\end{figure}

Overall, \evosops\ achieves fitness values of $F \geq 0.85$ (best genome) and $F \geq 0.73$ (final population mean) for all four collective behaviors, including near-optimal solutions ($F \geq 0.98$ best, $F \geq 0.95$ mean) to aggregation, phototaxing, and separation (see Table~\ref{tab:summary}).
\figtext~\ref{fig:fitnessdiversity} shows \evosops's fitness and population diversity progressions across the four collective behaviors.
Notably, independent \evosops\ runs produce similar fitness progressions despite having different random initial populations and SOPS executions.
For aggregation and phototaxing, fitness rapidly increases and then plateaus near optimality ($F \geq 0.95$) within the first 25 generations, followed by marginal improvements as population diversity decreases.
Separation follows a similar trend, but experiences an uptick in fitness ($F \geq 0.9$) when hypermutation is triggered by low population diversity at around generation 135.
Hypermutation was also enabled for object coating, but none of its runs met the population diversity threshold to trigger it.
This challenging new behavior saw decent but lower fitness values than the other three ($F = 0.85$ best, $F = 0.73$ mean) even after 750 generations.

Table~\ref{tab:summary} and \figtext~\ref{fig:fitnessdiversity} show that \evosops\ produces near-optimal algorithms that outperform their theoretical counterparts designed using the stochastic approach to SOPS.
Specifically, the best-fitness \evosops\ genome for aggregation achieves 4.21\% higher fitness than the~\citet{Li2021-programmingactive} algorithm for aggregation.
This improvement increases to 13.95\% higher fitness than the~\citet{Savoie2018-phototacticsupersmarticles} algorithm for phototaxing and again to 15.29\% higher fitness than the~\citet{Cannon2019-localstochastic} algorithm for separation.

Finally, \evosops\ discovers algorithms with desirable scalability to larger SOPS sizes and robustness to changes in their problem instances.
\figtext~\ref{fig:scaleruns} shows that the best-fitness genomes are similarly performant over three orders of magnitude of SOPS sizes, extending to SOPS that are 4x larger than those used in \evosops\ fitness evaluations.
\figtext~\ref{fig:bigconfigs} visualizes best-fitness executions on the largest SOPS sizes.
For aggregation, phototaxing, and separation, there is essentially no degradation of fitness at all; within the same $\numparticles^3$-step time horizon that was allowed in their fitness evaluations, these algorithms again achieve near-optimal fitness ($F \approx 0.98$).
The best-fitness genome for object coating is even more striking.
Despite its fitness evaluations consisting only of regular hexagonal objects and exactly enough particles to coat them with, this genome completely coats a large, irregular object and aggregates the extra particles nearby.
These results speak to the inherent scalability and robustness of local, memoryless algorithms and \evosops's ability to find high-quality solutions within that space.

\subsection{Diversity} \label{subsec:diversity}

In $\S$\nameref{subsec:efficacy}, we observed that independent \evosops\ runs for the same collective behavior displayed similar fitness progressions.
But \figtext~\ref{fig:diversity} shows that beneath this phenotypic similarity is a diverse exploration of genome space.
The PCA embeddings of all generations' genomes arrange independent runs as evenly distanced ``spokes'', confirming that each run explores distinct genome subspaces.
Clustering among the PCA embeddings reflects the varying convergence speeds shown in \figtext~\ref{fig:fitnessdiversity}.
Aggregation and phototaxing have sparse centers of initialization genomes and saturated, dense clusters of high-fitness genomes at the spokes' ends.
Separation and object coating, on the other hand, have much smoother trails and color gradients reflecting their steady fitness improvements over hundreds of generations.
Curiously, separation's hypermutation event is evident in both its PCA embeddings' unique staple-shaped turns and in its UMAP embeddings' visibly increased variance around each run's second bend.

The UMAP embeddings of these same genomes tell a complementary story about these behaviors' fitness landscapes.
Here, we see a stark contrast between aggregation and phototaxing's short jumps among scattered fitness niches and separation and object coating's long, winding journeys towards steadily higher fitness values.
Two of object coating's three runs even appear to loop back on themselves, perhaps struggling to find continued fitness improvements.
The shapes of these exploration trajectories thus provide a qualitative characterization of evolutionary search difficulty for each of these behaviors.

\subsection{Interpreting Genomes for New Algorithms} \label{subsec:interpret}

\begin{figure}[t]
    \centering
    \includegraphics[width=\columnwidth]{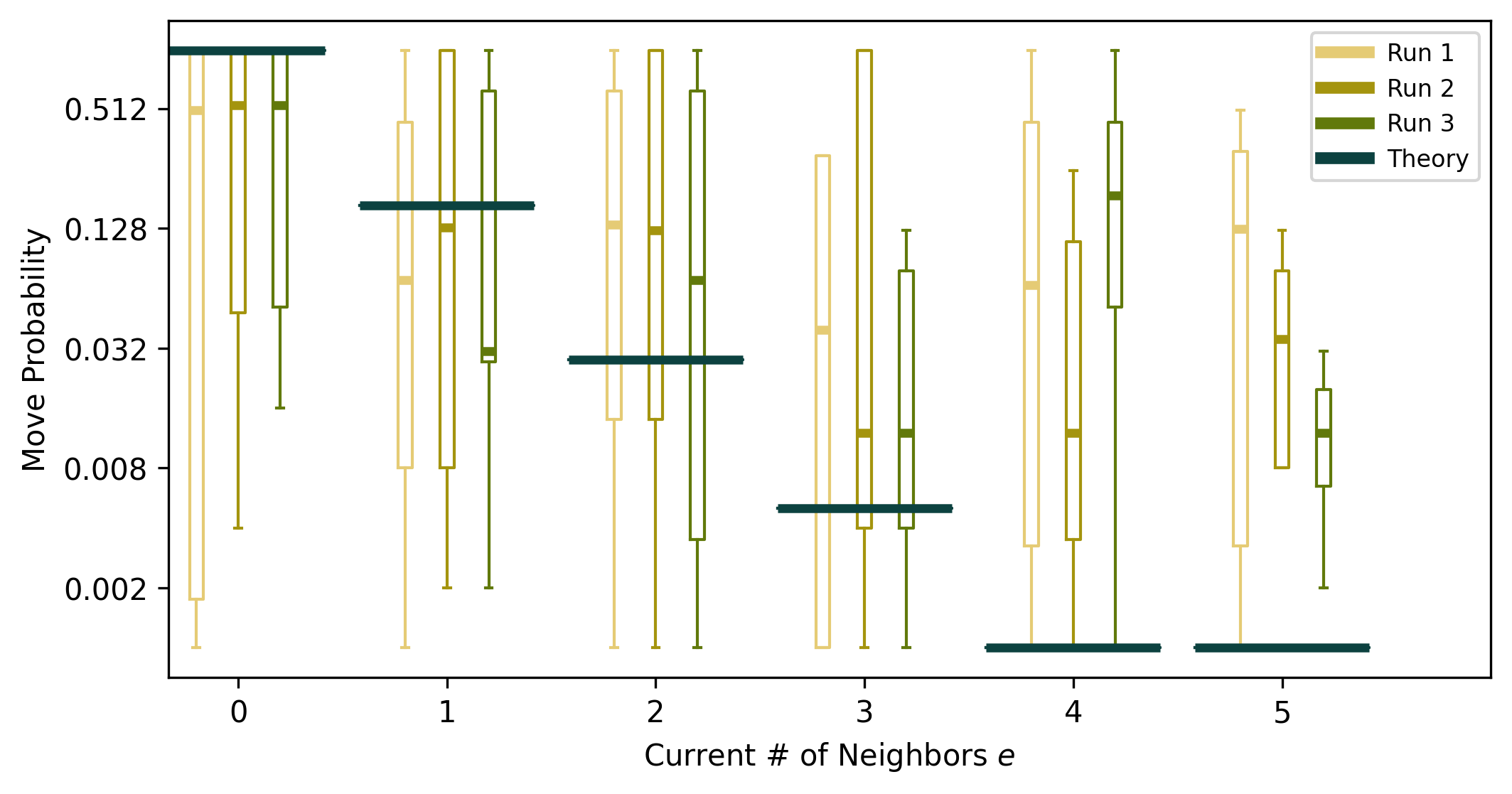}
    \vspace{-0.7cm}
    \caption{Gene-level comparison among the best-fitness aggregation genomes from three independent \evosops\ runs and the~\citet{Li2021-programmingactive} aggregation algorithm using $\lambda = 6$.
    We bin loci by a particle's current number of neighbors and plot the resulting distributions corresponding probabilities.}
    \label{fig:theorycomp}
\end{figure}

What local, memoryless interactions do high-fitness \evosops\ algorithms evolve to reliably drive their target behaviors?
As an illustrative example, we address this question for aggregation via a direct comparison to the stochastic approach to SOPS.
The~\citet{Li2021-programmingactive} aggregation algorithm defines a particle's move probability as $\lambda^{-e}$, where $\lambda > 1$ is a bias parameter affecting the strength of aggregation and $e$ is the particle's current number of neighbors.
In \figtext~\ref{fig:theorycomp}, we compare best-fitness aggregation genomes against these $\lambda^{-e}$ probabilities by binning loci sharing the same values of $e$ and plotting the corresponding move probability distributions.
For a fair comparison, we choose $\lambda = 6$ since Li et al.\ prove this value achieves aggregation.
The \evosops\ algorithms and stochastic approach both encourage particles to move when they have few neighbors ($e \leq 2$), but the theory strongly resists moves with $e \geq 3$ neighbors ($< 0.5\%$) whereas the \evosops\ algorithms are more permissive (1--15\%) and vary amongst themselves.
This suggests that unlike the stochastic approach's emphasis on contented particles with many neighbors, the most important factor for a successful aggregation algorithm may in fact be \textit{optimistic exploration in sparse neighborhoods}.
Taking this insight as inspiration, we developed a new and exceptionally simple algorithm---always move when $e \leq 2$, and otherwise move with 4\% probability---that performs as well as the Li et al.\ algorithm ($F = 0.95 \pm 0.011$, 100 trials, $\numparticles = 271$).

\section{Conclusion} \label{sec:conclude}

In this work, we presented \evosops, an evolutionary search framework for discovering stochastic, distributed algorithms for collective behavior in self-organizing particle systems.
With only a quality measure capturing the desired behavior, an (approximate) specification of an ideal configuration, and loci representing a particle's neighborhood information, \evosops\ effectively discovers local, memoryless algorithms that drive the desired behavior from the bottom up.
These algorithms are scalable, robust, and diverse and can outperform existing theoretical algorithms.
We also distilled the evolved aggregation algorithms into new design insights, suggesting how \evosops\ could aid future theoretical and practical investigations of computationally restricted agents and their collective behaviors.
Future work should explore how multiobjective and real-valued evolutionary search algorithms could enhance \evosops~\citep{Kennedy1995-particleswarm,Deb2002-fastelitist,Hansen2003-reducingtime}, characterize what collective behaviors \evosops\ can be successfully applied to, and develop a comprehensive methodology for gaining design insights from the high-fitness genomes \evosops\ produces.



\section*{Data \& Software Availability}

All source code and documentation supporting the results in this manuscript are openly available at
\ifanon
[URL omitted for blind review].
\else
\url{https://github.com/DaymudeLab/EvoSOPS}.
\fi

\ifanon\else
\section*{Acknowledgements}

We are grateful to Joseph Renzullo for his invaluable insights into enhancing population diversity.
The authors are supported in part by the ASU Biodesign Institute and the NSF under award CCF-2312537.
\fi

\footnotesize
\bibliographystyle{apalike}
\bibliography{ref}

\appendix

\end{document}